\documentclass[11pt,a4paper]{article}
\usepackage[hyperref]{acl2020}
\usepackage{times}
\usepackage{latexsym}
\usepackage{graphicx}
\usepackage{mathtools}
\usepackage{amsfonts}
\pdfoutput=1

\newcommand\Mycomb[2][^n]{\prescript{#1\mkern-0.5mu}{}C_{#2}}

\usepackage{microtype}

\aclfinalcopy % Uncomment this line for the final submission
\title{Multilogue-Net: A Context Aware RNN for Multi-modal Emotion Detection and Sentiment Analysis in Conversation}

\author{Aman Shenoy\thanks{* The following work was pursued when author was an intern at NVIDIA Graphics, Bengaluru} \\
  Birla Inst. of Technology and Science, Pilani \\
  Pilani, RA, India \\
  \texttt{f2016393@pilani.bits-pilani.ac.in} \\\And
  Ashish Sardana \\
  NVIDIA Graphics \\
  Bengaluru, KA, India\\
  \texttt{asardana@nvidia.com} \\}

\date{}

\begin{document}
\maketitle
\begin{abstract}
Sentiment Analysis and Emotion Detection in conversation is key in several real-world applications, with an increase in modalities available aiding a better understanding of the underlying emotions. Multi-modal Emotion Detection and Sentiment Analysis can be particularly useful, as applications will be able to use specific subsets of available modalities, as per the available data. Current systems dealing with Multi-modal functionality fail to leverage and capture - the context of the conversation through all modalities, the dependency between the listener(s) and speaker emotional states, and the relevance and relationship between the available modalities. In this paper, we propose an end to end RNN architecture that attempts to take into account all the mentioned drawbacks. Our proposed model, at the time of writing, out-performs the state of the art on a benchmark dataset on a variety of accuracy and regression metrics.
\end{abstract}

\section{Introduction}

Multi-modal Emotion Detection and Sentiment Analysis in conversation is gathering a lot of attention recently considering its potential use cases owing to the rapid growth of online social media platforms such as YouTube, Facebook, Instagram, Twitter etc. (\citealp{chen}, \citealp{poria}, \citealp{poria2}, \citealp{zadeh}, \citealp{zadeh2}), especially knowing that information obtained from any combination of more than one of the available modalities (e.g. text, audio, video) can be used to produce meaningful results.  

The current state of the art systems on multi-modal emotion detection and sentiment analysis do not treat the modalities in accordance to the information they are capable of holding (e.g. textual information is significantly more likely to hold contextual information then audio or video features are), lack an adequate fusion mechanism, and fail to effectively capture the context of a conversation in a multi-modal setting. In addition to the lack of proper usage of the available modalities, models also fail to effectively capture the flow of a conversation, the separation between speaker and listener states, and the emotional effect a speaker’s utterance has on the listener (s) in dyadic conversations.  

Our proposed model Multilogue-Net, attempts to embed basic domain knowledge and takes insight from \citet{poria3}, assuming that the sentiment or emotion governing a particular utterance predominantly depends on 4 factors – 	interlocutor state, interlocutor intent, the preceding and future emotions, and the context of the conversation. Interlocutor intent amongst the mentioned is particularly difficult to model due to its dependency of prior knowledge about the speaker, but modelling the other 3 separately, yet in an interrelated manner was theorized to produce meaningful results if managed to be captured effectively. The key intention was to attempt to simulate the setting in which an utterance is said, and use the actual utterance at that point to be able to gain better insights regarding emotion and sentiment of that utterance. The model uses information from all modalities learning multiple state vectors (representing interlocutor state) for a given utterance, followed by a pairwise attention mechanism inspired by \citet{ghosal}, attempting to better capture the relationship between all pairs of the available modalities.	

The model uses two gated recurrent units (GRU) \cite{chung} for each modality for modelling interlocutor state and emotion. Along with these GRU's, the model also uses an inter-connected context network, consisting of the same number of GRU's as the number of available modalities, to model a different learned context representation for each modality. The incoming utterance representations and the historical GRU outputs are used at every timestamp to be able to arrive at a prediction for that timestamp. 

The model produces $m$ different representations at every timestamp (Where $m$ is the number of modalities), where each representation is the emotional state at that timestamp as conveyed by each of the modalities. These $m$ representations are used by the fusion mechanism to incorporate information from each of the $m$ representations to be able to arrive at the final prediction for that timestamp. We understand that the usage of the pairwise attention mechanism, along with the Emotion GRU are what make the model flexible across tasks.

The usage of only the text representation as input to the context GRU’s has been observed to be key to the results, as the context of the conversation would be better captured by textual information then it would have with audio or video information. We believe that Multilogue-net performs better than the current state of the art \cite{ghosal} on multi-modal datasets because of better context representation leveraging all available modalities.\footnote{A basic model and training implementation of Multilogue-Net can be found at \url{https://github.com/amanshenoy/multilogue-net}.}

The remaining sections of the paper are arranged as follows: Section 2 – discusses related work; Section 3 – discusses the model in detail; Section 4 – provides experimental results, dataset details, and analysis; Section 5 contains our ablation studies and its implications; and finally Section 6 – speaks on potential future work, and concludes our paper.

\section{Related Work}

Multi-modal Emotion recognition and Sentiment Analysis has always attracted attention in multiple fields such as natural language processing, psychology, cognitive science, and so on \cite{picard}. Previous works have been done studying factors of variation that have a more direct correlation with emotion, such as \citet{ekman}, who found correlation between emotion and facial cues, and a lot of studies extensively focus on emotions and their relationship with one another such as Plutchik’s wheel of emotions, which defines eight primary emotion types, each of which has a multitude of emotions as sub-types.

Early work done to leverage multi-modal information for emotion recognition includes works such as \citet{datcu}, who fused acoustic information with visual cues for emotion recognition and \citet{opensmile}, who used contextual information for emotion recognition in multi-modal settings. More recently, deep recurrent neural networks have been used to be able make the best of the learned representations of the modalities available to be able to give very effective and accurate emotion and sentiment predictions. \citet{poria2} successfully used RNN-based deep networks for multi-modal emotion recognition, which was followed by multiple other works (\citealp{chen}; \citealp{mfn}; \citealp{zadeh3}) giving results far better than what was seen before. Recent works also include works such as \citet{hazarika}, who used memory networks for emotion recognition in dyadic conversations, where two distinct memory networks enabled inter-speaker interaction. 

Some works such as DialogueRNN \cite{dialoguernn}, though focused on emotion recognition and sentiment analysis using a single modality (text), works very well in a multi-modal setting by just replacing the text representation with a concatenated vector of all the modality representations. DialogueRNN effectively leveraged the separation between the speakers by maintaining two independent gated recurrent units to keep track of the interlocutor states, also effectively capturing context in the conversation, yielding state-of-the-art performance on uni-modal data. Even though DialogueRNN was able to give reasonably good results on multi-modal data, the lack of an adequate fusion mechanism and the lack of focus on a multi-modal representation held its multi-modal performance back. 

Apart from the kind of works shown before, where a methodology or a model was proposed, works such as \citet{poria3} spoke extensively about the research challenges and advancements in emotion detection in conversation and gave a comprehensive overview of the problem. Most recently \citet{ghosal} introduced the idea of learning the relationship between pairs of all available modalities using pairwise attention, in a multi-modal setting, where similar attributes learned by multiple modalities are emphasized and differences between the modality representations are diminished. Pairwise attention proved to be incredibly effective yielding state-of-the-art performance on multi-modal data with just simple representations for each modality.

\section{Proposed Methodology}
\subsection{Problem Formulation}
Let there be a $P$ number of participants $p_1, p_2, ... , p_P$  in the conversation. The problem is defined such that for every utterance $u_1, u_2, ..., u_N$  uttered by any participant(s), a sentiment score is allotted along with a predicted emotion label (one of happy, sad, angry, surprise, disgust, and fear). Each utterance corresponds to a particular participant of the conversation, allowing this formulation of the problem to also capture the average sentiment of a participant in the conversation. Predictions over utterances also avoid problems such as classification during long moments of silence when predictions are made for a fixed time interval, and is also mostly common practice. 

For every utterance $u_{t}(p)$, where $p$ is the party who uttered the utterance, there exist three independent representations , $t_t \in \mathbb{R}^{D_t}$ , $a_t \in \mathbb{R}^{D_a}$, and $v_t \in  \mathbb{R}^{D_v}$, and are obtained using the feature extractors further explained in section 4.2.

This gives us our overall formulation of the problem, which is to be able to learn a function which would take as input three independent representations of a particular utterance, information regarding the previous emotional state of the participant, and a representation of the current context of the conversation - to be able to map to an output prediction of a sentiment score and emotion label. 

Details regarding how these representations are updated and how the output is generated using these inputs are described in detail below.

\subsection{Model Details}

Modelling was done under the underlying assumption that the sentiment or emotion of an utterance predominantly depends on four factors as mentioned before:
\begin{itemize}
    \item{Interlocutor State}
    \item{Interlocutor Intent}
    \item{Context of the conversation until that point}
    \item{Previous interlocutor states and emotions of a particular participant in the conversation}
\end{itemize}

The proposed model attempts to model three out of the mentioned four explicitly, and assume that interlocutor intent will be modelled implicitly during model training. Interlocutor state is modelled using a state GRU (will be referred to as $sGRU$), A context GRU is used to keep track of the context of the conversation ($cGRU$), and an emotion GRU ($eGRU$) is used to keep track of the emotional state of that particular participant. Finally, a pairwise attention mechanism, which uses the emotion representation of all modalities at a particular timestamp is used to leverage the important modalities and relevant combination of the modalities for emotion or sentiment prediction at that timestamp.

\begin{figure}[htbp]
	\centering
		\includegraphics[width=220pt]{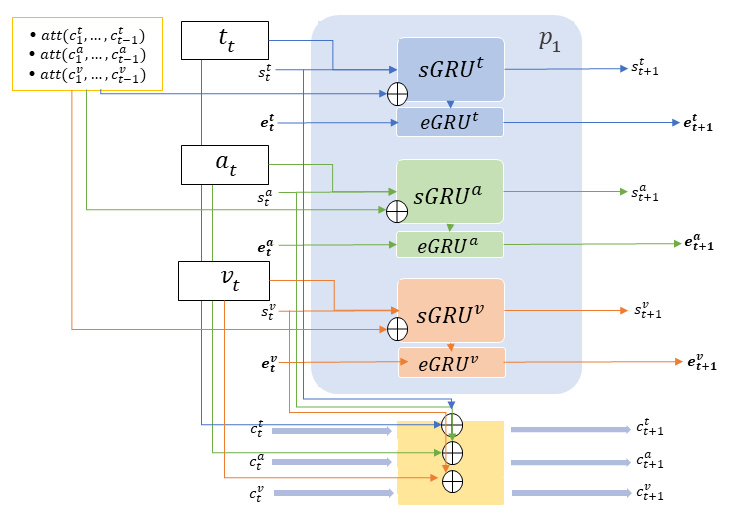}    
    \caption[monologue]{Description of all the state updates at timestamp $t$ for a single participant $p_1$}
	\label{fig:monologue}
\end{figure}

Every utterance has three independent feature representations (text, audio, and video features), $t_t \in \mathbb{R}^{D_t}$ , $a_t \in \mathbb{R}^{D_a}$, and $v_t \in  \mathbb{R}^{D_v}$. Each of these feature representations are treated and operated on independently until the pairwise attention mechanism. The model consists of two GRU’s (state GRU, and emotion GRU) for every modality and participant, and a context GRU for each modality common to all participants in the conversation (If $p$ is the number of participants and $m$ is the number of modalities, the model would have a total of $2mp + m$ GRU’s). The inputs at the current timestamp and the previous state, context, and emotion representations are operated on to be able to arrive at the prediction at that timestamp. Figure \ref{fig:monologue} describes the updates at a particular timestamp and the role of each GRU is further explained below.

\begin{figure*}[htbp]
	\centering
		\includegraphics[width=440pt]{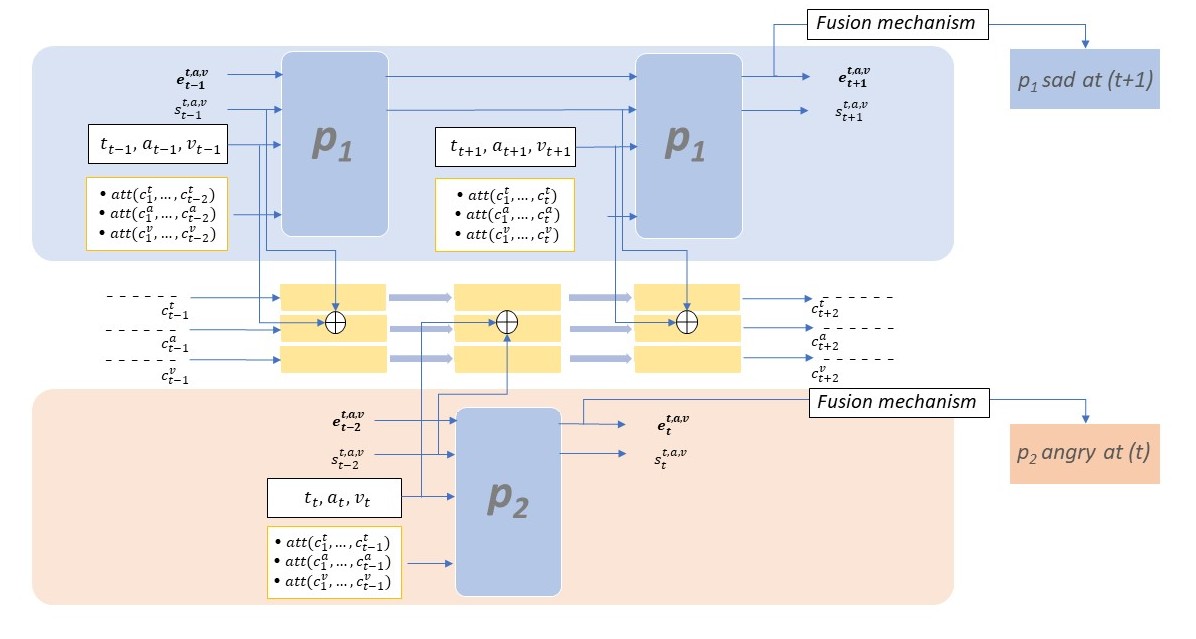}    
    \caption[attention]{State updates and final prediction output in a conversation between two participants $p_1$ and $p_2$, where the updates of each participant at a timestamp is as given in figure \ref{fig:monologue}}
	\label{fig:dialogue}
\end{figure*}

\subsubsection{Context GRU ($cGRU$)}
The Context GRU ($cGRU$) for each modality aims to capture the context of the conversation by jointly encoding the utterance representation of that modality (at timestamp $t$ in the given diagram) ($t_t \in \mathbb{R}^{D_t}$ , $a_t \in \mathbb{R}^{D_a}$, or $v_t \in  \mathbb{R}^{D_v}$) and the previous timestamp speaker state GRU output of that modality. This accounts for inter-speaker and inter-utterance dependencies to produce an effective context representation. The current utterance $t_t$, $a_t$, or $v_t$,  changes the state of that speaker from ($s_t^t$, $s_t^a$, $s_t^v$) to ($s_{t+1}^t$, $s_{t+1}^a$, $s_{t+1}^v$). To capture this change in context we use GRU cell $cGRU$ having output size $D_c$, using $t_t$, $a_t$, or $v_t$ and ($s_t^t$, $s_t^a$, $s_t^v$) as:

\begin{equation}\label{eq:1}
    c_{t+1}^t = cGRU(c_t^t, (t_t \oplus s_t^t))  
\end{equation}

\begin{equation}\label{eq:2}
    c_{t+1}^a = cGRU(c_t^a, (a_t \oplus s_t^a)) 
\end{equation}

\begin{equation}\label{eq:3}
    c_{t+1}^v = cGRU(c_t^v, (v_t \oplus s_t^v))  
\end{equation}

Where $D_c$ is the size of the context vectors $c_{t+1}^t$, $c_{t+1}^a$, and $c_{t+1}^v$.$D_t$, $D_a$, and $D_v$ are the sizes of utterance representations of text, audio, and video respectively.$\oplus$ represents the concatenation operation, $D_s$ is the size of all the state vectors $s_{t+1}^t$, $s_{t+1}^a$, and $s_{t+1}^v$; and all GRU weight and biases shapes are such that they produce the expected shape of outputs taking the given shape of inputs.  

\subsubsection{State GRU ($sGRU$)}
The network keeps track of the participants involved in a conversation by employing a $p*m$ number of ($sGRU$)'s, where $p$ is the number participants in the conversation and $m$ is the number of available modalities.The $sGRU$ associated with a participant outputs fixed size vectors which serve as an encoding to represent the interlocutor state, and are directly used for both emotion and sentiment prediction, and updating the context vectors.

All the state vectors are initialized to null at the first timestamp. For a timestamp $t$, the state vector of participant $p$ and modality $m \in \{t, a, v\}$ is updated using the input feature representation of that modality and simple attention over all the context vectors until that timestamp. The simple attention mechanism over all the context vectors is described by the following equations:

\begin{equation}\label{eq:4}
    \alpha = softmax({m_t^T}{W_{\alpha}}{[c_1^m, c_2^m, ..., c_t^m]})
\end{equation}
\begin{equation}\label{eq:5}
    att_t = {\alpha}{[c_1^m, c_2^m, ..., c_t^m]^T}  
\end{equation}

Where $m_t^T \in \{t_t^T, a_t^T, v_t^T\}$, $W_{\alpha} \in \mathbb{R}^{D_{t, a, v} \times D_c}$, ${\alpha}^T \in \mathbb{R}^{(t-1)}$, and $att_t \in \mathbb{R}^{D_c}$. In equation \ref{eq:4}, we calculate attention scores over all previous context representations of all previous utterances, highlighting the relative importance of all the previous context vectors to $m_t$. A softmax layer is applied to amplify this relative importance, and finally equation \ref{eq:5} the final output of attention over context $att_t$ is calculated by pooling the previous context vectors with $\alpha$.

We then employ $sGRU^{t, a, v}$ to update $s_t^{t, a, v}$ to $s_{t+1}^{t, a, v}$ on the basis of incoming utterance representations for each modality $m_t^T \in \{t_t^T, a_t^T, v_t^T\}$ and the context representations $att_t^t$, $att_t^a$, and $att_t^v$ using GRU cells $sGRU_t^t$, $sGRU_t^a$, and $sGRU_t^v$, each of output size $D_s$.

\begin{equation}\label{eq:6}
    s_{t+1}^t = sGRU(s_t^t, (t_t \oplus att_{t+1}^t))  
\end{equation}

\begin{equation}\label{eq:7}
    s_{t+1}^a = sGRU(s_t^a, (a_t \oplus att_{t+1}^a)) 
\end{equation}

\begin{equation}\label{eq:8}
    s_{t+1}^v = sGRU(s_t^v, (v_t \oplus att_{t+1}^v))  
\end{equation}

Where $D_s$ is the size of all the state vectors $s_{t+1}^t$, $s_{t+1}^a$, and $s_{t+1}^v$.$D_t, D_a, D_v$ are the sizes of utterance representations of text, audio, and video respectively.$\oplus$ represents concatenation operation, and all GRU weights shapes are such that they produce the expected shape of outputs taking the given shape of inputs.

The intended purpose of using this as the input to $sGRU^{t, a, v}$ is to model the dependency of the speaker state on the context of the conversation as understood by the utterances until that point, along with the utterance representation at that point. The output of the $sGRU$ for modality $m$ and timestamp $t$ serves as an encoding of the speaker state as conveyed by modality $m$, at time $t$.

\subsubsection{Emotion GRU ($eGRU$)}

The emotion GRU serves as the decoder for the encoding produced by the state GRU. The emotion GRU uses the previous timestamp $eGRU$ output, and the encoding provided by $sGRU$ to produce an emotion or sentiment representation which is further used by the pairwise attention mechanism to be able to produce the relevant output for prediction. At timestamp $(t+1)$ the emotion vectors are updated as:

\begin{equation}\label{eq:9}
    e_{t+1}^t = eGRU(e_t^t, s_{t+1}^t)  
\end{equation}

\begin{equation}\label{eq:10}
    e_{t+1}^a = eGRU(e_t^a, s_{t+1}^a) 
\end{equation}

\begin{equation}\label{eq:11}
    e_{t+1}^v = eGRU(e_t^v, s_{t+1}^v)  
\end{equation}

Where $D_e$ is the size of all the emotion vectors $e_{t+1}^t$, $e_{t+1}^a$, and $e_{t+1}^v$.$D_t, D_a, and D_v$ are the sizes of utterance representations of text, audio, and video respectively.$D_e$ is the size of the state vectors $s_{t+1}^t$, $s_{t+1}^a$, and $s_{t+1}^v$; and all GRU weights shapes are such that they produce the expected shape of outputs taking the given shape of inputs. 

The emotion GRU acts as a decoder to the encoding produced by the associated state GRU, producing a vector which can be used for both sentiment and emotion prediction.

\subsubsection{Pairwise Attention Mechanism}

The emotion GRU for each timestamp will produce an $m$ number of vectors (where $m$ is the number of modalities available). Pairwise attention is then used over these $m$ vectors to produce the final prediction output. In particular pairwise attention is calculated over the following pairs in our case – $(e^v, e^t), (e^t, e^a)$, and $(e^a, e^v)$. Pairwise attention for pair $(e^v, e^t)$ would be calculated as follows:

\begin{figure}[htbp]
	\centering
		\includegraphics[width=205pt, angle=90]{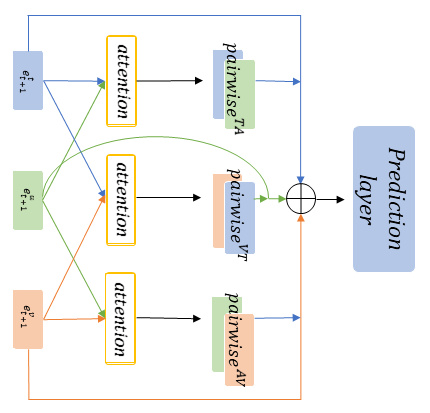}    
    \caption[fusion]{Pairwise attention mechanism used as the fusion mechanism followed by the final prediction layer}
	\label{fig:fusion}
\end{figure}

\begin{equation}\label{eq:12}
    B_1 = e^v.(e^t)^T, B_2 = e^t.(e^v)^T   
\end{equation}
\begin{equation}\label{eq:13}
    N_1 = softmax(B_1), 
    N_2 = softmax(B_2) 
\end{equation}
\begin{equation}\label{eq:14}
    O_1 = N_1.e^t, O_2 = N_2.e^v
\end{equation}
\begin{equation}\label{eq:15}
    A_1 = O_1 \odot e^v, A_2 = O_2 \odot e^t
\end{equation}
\begin{equation}\label{eq:16}
    pairwise(e^v, e^t) = A_1 \oplus A_2
\end{equation}

Where $B_1, B_2 \in \mathbb{R}^{D_e \times D_e}$; $N_1, N_2 \in \mathbb{R}^{D_e \times D_e}$; $A_1, A_2 \in \mathbb{R}^{D_e \times D_e}$; and $pairwise(e^v, e^t) \in \mathbb{R}^{D_e \times 2D_e}$; $\odot$  represents element-wise product; and $\oplus$ represents concatenation.

A complete analysis on the pairwise attention mechanism has been done by \citet{ghosal}, where the role of each one of the intermediate variables has been described. These equations (\ref{eq:12}, \ref{eq:13}, \ref{eq:14}, \ref{eq:15}, \ref{eq:16}) calculate $\Mycomb[m]{2}$ pairwise fusion representations, which are further concatenated to make the final prediction as described below.

\subsubsection{Final Predictions}
The prediction layer varies based on whether a sentiment or emotion prediction is expected. For sentiment prediction first all three pairs of pairwise attention i.e. $pairwise(e^v, e^t)$, $pairwise(e^a, e^t)$, and $pairwise(e^v, e^a)$ at that timestamp are concatenated along with the emotion GRU outputs at that timestamp ($e_t^t$, $e_t^a$, and $e_t^v$) and the concatenated layer is passed through a fully connected layer followed by a $softmax$ or $tanh$ layer based on the nature of the expected prediction. For sentiment prediction between -1 and +1 at timestamp t the output layer would equate as follows:

\begin{equation}\label{eq:17}
    pw = pw(e^v, e^t) \oplus pw(e^a, e^t) \oplus pw(e^v, e^a)
\end{equation}
\begin{equation}\label{eq:18}
    L_t = pw \oplus e_t^t \oplus e_t^a \oplus e_t^v
\end{equation}
\begin{equation}\label{eq:19}
    pred_{sentiment(t)} = tanh({W_L}{L_t})
\end{equation}

Where $pairwise(e^v, e^t)$ has been represented as $pw(e^v, e^t)$; and $W_L \in \mathbb{R}^{9D_e \times 1}$.

For emotion prediction we use a fully connected layer along with a final $softmax$ layer to calculate 6 emotion class probabilities from $L_t$.

\begin{equation}\label{eq:20}
    l_t = ReLU({W_l}{L_t} + {b_l})
\end{equation}
\begin{equation}\label{eq:21}
    P_t = softmax({W_{smax}}{l_t} + b_{smax})
\end{equation}
\begin{equation}\label{eq:22}
    pred_{emotion(t)} = argmax({P_t})
\end{equation}

Where $W_l \in \mathbb{R}^{D_l \times 9D_e}$; $b_l = \in \mathbb{R}^{D_l}; {W_{smax}} \in \mathbb{R}^{c \times D_l}; b_{smax} \in \mathbb{R}^c$ and $P_t \in \mathbb{R}^c$

\subsubsection{Training}
Fairly standard practices have been employed for the training of the model. Categorical cross-entropy has been used along with L2-regularization as the loss function during training for emotion prediction, to maximize likelihood over each of the classes.

Mean Square Error (MSE) along with L2 regularization has been employed as loss function during training for sentiment regression. The usage of a saturating output layer and a loss function that does not undo the saturation, leads to the model to stop training when it makes extreme predictions (close to -1 or +1) due to very small gradients. Using initialization strategies that start at smaller model weights, mini-batch gradient descent-based Adam \cite{adam} optimizer, and using L2 regularization is used to avoid this failure mode.

\section{Experiments, Datasets, and Results}

\begin{table}
\centering
\begin{tabular}{lll}
\hline
\textbf{Metric} & \textbf{A2} & \textbf{F1}\\
\hline
\textbf{Text + Audio} \\
\hline
BC-LSTM & 79.30 & - \\
MMMU-BA & \textbf{80.58} & - \\
DialogueRNN & 78.81 & \textbf{79.12} \\
Multilogue-net & 80.12 & 78.84 \\
\hline

\textbf{Video + Audio} \\
\hline
BC-LSTM & 62.10 & - \\
MMMU-BA & 65.16 & - \\
DialogueRNN & 63.22 & 60.14 \\
Multilogue-net & \textbf{69.55} & \textbf{63.40} \\
\hline

\textbf{Text + Video} \\
\hline
BC-LSTM & 80.20 & - \\
MMMU-BA & \textbf{81.51} & - \\
DialogueRNN & 79.88 & 79.10 \\
Multilogue-net & 80.66 & \textbf{79.62} \\
\hline

\textbf{Text + Audio + Video} \\
\hline
BC-LSTM & 80.30 & - \\
MMMU-BA & \textbf{82.31} & - \\
DialogueRNN & 79.80 & 79.48 \\
Multilogue-net & 81.19 & \textbf{80.10} \\
\hline

\end{tabular}
\caption{\label{tab:mosi}
Multilogue-Net performance on CMU-MOSI in comparison with the current and previous state-of-the-art on the dataset. A2 indicating accuracy with 2 classes, and F1 indicating F1 score .
}
\end{table}

\subsection{Datasets}
We evaluate our model using two benchmark datasets - CMU Multi-modal Opinion-level Sentiment Intensity (CMU-MOSI) \cite{mosi}  and the recently published CMU Multi-modal Opinion Sentiment and Emotion Intensity (CMU-MOSEI) dataset \cite{mosei}.

\subsubsection{CMU-MOSI}
CMU-MOSI dataset consists of 93 videos spanning over 2199 utterances. Each utterance has a sentiment label associated with it. It has 52, 10 \& 31 videos in training, validation \& test set accounting for 1151, 296 \& 752 utterances. CMU-MOSEI has 3229 videos with 22676 utterances from more than 1000 online YouTube speakers. The training, validation \& test set consist of 16216, 1835 \& 4625 utterances, respectively. Each utterance in CMU-MOSI dataset has been annotated as either positive or negative. 

\subsubsection{CMU-MOSEI}
In CMU-MOSEI dataset labels are in a continuous range of -3 to +3 and are accompanied by an emotion label being one of six emotions. However, in this work we also project the instances of CMU-MOSEI in a two-class classification setup with values $\geq$ 0 signifies positive sentiments and values $<$ 0 signify negative sentiments. We have called this A2 accuracy (accuracy with 2 classes). Along with this we have also shown results for continuous range prediction between -3 and +3, and emotion prediction with the 6 emotion labels for each utterance in CMU- MOSEI. We have used A2 as a metric to be consistent with the previous published works on CMU-MOSEI dataset (\citealp{ghosal}; \citealp{mosei}). CMU-MOSEI has further been used for other comprehensive experiments due to its large sizer and easier feature extraction
\subsection{Uni-modal Feature Extraction}

\begin{table}
\centering
\begin{tabular}{lllll}
\hline
\textbf{Metric} & \textbf{A2} & \textbf{F1}& \textbf{MAE} & \textbf{r}\\
\hline
\textbf{T + A} \\
\hline
MMMU-BA & 79.74 & - & - & - \\
DialogueRNN & 79.80 & 78.32 & - & -\\
Multilogue-net & \textbf{80.18} & \textbf{79.88} & - & -\\
\hline

\textbf{V + A} \\
\hline
MMMU-BA & \textbf{76.66} & - & - & - \\
DialogueRNN & 73.90 & 73.92 & - & -\\
Multilogue-net & 75.16 & \textbf{74.04} & - & -\\
\hline

\textbf{V + T} \\
\hline
MMMU-BA & 79.40 & - & - & - \\
DialogueRNN & 78.90 & 78.12 & - & -\\
Multilogue-net & \textbf{80.06} & \textbf{79.84} & - & -\\
\hline

\textbf{T + A + V} \\
\hline
Graph-MFN  & 76.90 & 77.00 & 0.71 & \textbf{0.54} \\
MMMU-BA & 79.80 & - & - & - \\
DialogueRNN & 79.98 & 79.82 & 0.69 & 0.42\\
Multilogue-net & \textbf{82.10} & \textbf{80.01} & \textbf{0.59} & 0.50\\
\hline

\end{tabular}
\caption{\label{tab:mosei}
Multilogue-Net performance on CMU-MOSEI Sentiment Labels compared to previous state-of-the-art models on regression and accuracy Metrics. All metrics apart from MAE represents higher values for better results, MAE represents lower values for better results.
}
\end{table}

\subsubsection{CMU-MOSEI}
We use the CMU-Multi-modal Data SDK \cite{mosei} for feature extraction. For MOSEI dataset, sentiment label-level features were provided where text features used were GloVe embeddings \cite{glove}, visual features extracted by Facet \cite{facet} \& acoustic features by OpenSMILE \cite{opensmile}. Thereafter, we compute the average of sentiment label-level features in an utterance to obtain the utterance-level features. For each sentiment label-level feature, the dimension of the feature vector is set to 300 (text), 35 (visual) \& 384 (acoustic).

\subsubsection{CMU-MOSI}
In contrast, for MOSI dataset we use utterance level features provided in \citet{poria2}. These utterance-level features represent the outputs of a convolutional neural network \cite{cnn}, 3D convolutional neural network \cite{3dcnn} \& openSMILE \cite{opensmile} for text, visual \& acoustic modalities, respectively. Dimensions of utterance-level features are 100, 100 \& 73 for text, visual \& acoustic, respectively.

\subsection{Experiments}

We evaluate our proposed approach on CMU-MOSI (test-set) on accuracy and F1 score, and CMU-MOSEI (dev-set) on accuracy, F1 score, mean absolute error ($MAE$), pearson score ($r$), and accuracy's on the emotion labels. Due to the lack of speaker information in CMU-MOSI we were not able to use the CMU-Multi-modal Data SDK for sentiment label extraction, to be able to evaluate our approach on CMU-MOSI on mean absolute error and Pearson score.

\begin{table*}
\centering
\begin{tabular}{|c|c|c|c|c|c|c|c|c|c|c|c|c|}
\hline
\multicolumn{13}{|c|}{\textbf{MOSEI Emotions (Text + Video + Audio)}} \\
\hline
\textbf{Emotion} & \multicolumn{2}{|c|}{Anger} & \multicolumn{2}{|c|}{Disgust} & \multicolumn{2}{|c|}{Fear} & \multicolumn{2}{|c|}{Happy} & \multicolumn{2}{|c|}{Sad} & \multicolumn{2}{|c|}{Surprise} \\
\hline
Metric & WA & F1 & WA & F1 & WA & F1 & WA & F1 & WA & F1 & WA & F1 \\
\hline
Graph-MFN  & 62.6 & 72.8 & 69.1 & 76.6 & 62.0 & \textbf{89.9} & 66.3 & 66.3 & 60.4 & 66.9 & 53.7 & \textbf{85.5} \\
Multilogue-Net  & \textbf{83.1} & \textbf{80.9} & \textbf{90.3} & \textbf{87.3} & \textbf{89.7} & 87.0 & \textbf{70.0} & \textbf{68.4} & \textbf{76.1} & \textbf{74.5} & \textbf{87.4} & 84.0 \\
\hline

\end{tabular}
\caption{\label{tab:mosei2}
Multilogue-Net performance on MOSEI Emotion Labels compared with that of Graph-MFN on weighted accuracy and F1 score. MOSEI Emotion label results were presented by only one model, and comprehensive results have not been published for the same.
}
\end{table*}
Results have also been reported for usage of two of the three available modalities. Uni-modal performance has not been reported as the focus of the paper is the effective usage of multi-modal data. In a uni-modal setting the model would not be using the fusion mechanism and the output would be equivalent to having a few dense layers after the emotion GRU to directly output the final prediction. F1 scores have not been mentioned by most previous models being used for comparison, but have been reported for Multilogue-Net for additional comparison to any future models using CMU-MOSI dataset.

Table \ref{tab:mosi} shows the performance of Multilogue-Net on CMU-MOSI dataset, comparing to the current state of the art \cite{ghosal}, previous state-of-the-art \cite{poria2}, and DialogueRNN \cite{dialoguernn} (Multi-modal performance of DialogueRNN has not been reported by \citet{dialoguernn}, and we have run these experiments additionally for a better comparative study, where concatenating the input representations has been used as a fusion mechanism). Our model consistently outperforms the previous state-of-the-art but performs better only on one of the subsets of the modalities when compared to the current state-of-the-art. 

In comparison to MMMU-BA our model also lacks in Multi-modal performance. We theorize that the model performance is lacking because of the low number of training examples (CMU-MOSI consists only of 93 conversations out of which 62 were used for training), in contrast to our model which has a high capacity (Relative to models being compared with). Since Multilogue-Net learns a lot of intermediate representations in order to make a prediction, it would need a larger dataset with more variability to be able to learn meaningful representations. The proposition that performance lacks due to a lack of training examples is backed by the results on CMU-MOSEI (demonstrated in a comparative setting in Table \ref{tab:mosei} and \ref{tab:mosei2}) where the model consistently outperforms the current state-of-the-art on most metrics. 

On CMU-MOSEI, our model seems to perform very consistently on both sentiment and emotion labels. The model outperforms the current state of the art on all but one metric (both classification and accuracy) on sentiment labels in the tri-modal setting. Multilogue-Net also outperforms the current state of the art on the emotion labels by a considerable margin (This is also attributed to the fact that not a lot of models have presented results on these labels).

Similar observations are made in both datasets, where the tri-modal metrics show the best performance, and audio + video show the worst relative performance (suggesting the importance of text in a multi-modal setting). Textual information seems to be the guiding factor for multi-modal performance, with video and audio features simply acting as a push to the uni-modal performance on text. 

We theorize that the performance of Multilogue-Net is majorly attributed to its increased capacity as compared to previous models. Effective usage of this increased capacity, using representations inspired from a basic understanding of conversation, along with a larger dataset for training have been key in achieving the improved results.

\section{Ablation Studies and Analysis}
Until now, some architectural considerations, such as the use of $eGRU$ and the fusion mechanism, have been briefly explained but not empirically justified. This section aims to get empirical evidence regarding the effectiveness of these modules. Since our model completely hinges around the usage of the context and state GRU's, our ablation studies and analysis have focused on the fusion mechanism and emotion GRU ($eGRU$) only. 

\subsection{Fusion Mechanism}
The effectiveness of the fusion mechanism can be very easily examined by observing the results of the model on both tasks $-$ Sentiment Regression and Emotion Recognition, with and without the fusion mechanism. Table \ref{tab:fusion} shows these results on CMU-MOSEI modality subsets.

\begin{table}
\centering
\begin{tabular}{lll}
\hline
\textbf{Fusion Mechanism} & \textbf{A2} & \textbf{MAE}\\
\hline
\textbf{Text + Audio} \\
\hline
without & 75.78 & - \\
with & \textbf{80.18} & - \\
\hline
\textbf{Video + Audio} \\
\hline
without & \textbf{75.66} & - \\
with & 75.16 & - \\
\hline
\textbf{Text + Video} \\
\hline
without & 76.80 & - \\
with & \textbf{80.06} & - \\
\hline
\textbf{Text + Audio + Video} \\
\hline
without & 79.80 & 0.66 \\
with & \textbf{82.10} & \textbf{0.59} \\
\hline

\end{tabular}
\caption{\label{tab:fusion}
Multilogue-Net performance on CMU-MOSEI with and without the fusion mechanism - for 'without' fusion we have concatenated all the representations and directly passed them to the prediction layer.
}
\end{table}

The bi-modal results in table \ref{tab:fusion} involve evaluating the pairwise attention module only once (Since there is only one pair available), directly followed by the prediction layer. The tri-modal case on the other hand involves evaluating the pairwise attention module thrice (Once for each pair). In general, the number of times this module will have to be evaluated for $m$ modalities is $\Mycomb[m]{2}$, which raises a fair concern regarding the trade-off between the additional computational cost and performance. 

We empirically observe that the additional computational cost can be considered negligible in context of the increased performance, largely attributing to the non-parametric nature of the fusion mechanism and the relatively small number of additional parameters in the prediction layer ($6D_e$ for the sentiment regression; $36D_e$ for emotion recognition).

The fusion mechanism seems to clearly be beneficial in all of the reported cases apart from video + audio, implying that the fusion mechanism is useful only in the cases the text representation is used. This further strengthens our claim that the text representation guides tri-modal performance.

\subsection{Emotion GRU ($eGRU$)}

Unlike as done with the fusion mechanism, the effectiveness of the $eGRU$ cannot be examined by evaluating metrics with and without it. Removing the Emotion GRU would clearly be detrimental to the results, and would not convey the intention of having it.

The primary intention of having the $eGRU$ can be considered to be maintaining consistency between tasks. To better understand what this means table \ref{tab:geru} quantitatively demonstrates this effect. The model was trained separately for Emotion Detection and Sentiment Regression tasks. After both the models were trained satisfactorily, a particular sample from the test set (test sample 6) was inferred on. We then retrieved the intermediate text representations ($e_4^t$, $c_4^t$, and $s_4^t$; superscript $t$ indicating text modality) at a particular timestamp ($t = 4$) for both models on that sample. The Euclidean Distance between these two sets of representations (one for each task) was evaluated and have been shown in table \ref{tab:geru}, where we can clearly observe that the euclidean distance between the emotion representations is much larger as compared to the state and context representations. 

This shows that for both tasks, interlocutor state and context representations are relatively similar to each other, whereas the emotion state representation is more varied and task dependant. This not only allows us to use the same $cGRU$ and $sGRU$ weights across tasks, but would also allow us to train for multiple tasks in parallel using a different $eGRU$ for each task - giving us consistent and accurate predictions across multiple tasks. Analysis of such a network, and whether training for multiple tasks in parallel aids one another, has not been covered in this paper and is left to our future work.

\begin{table}
\centering
\begin{tabular}{ll}
\hline
\textbf{Representation} & \textbf{Euclidean Distance} \\
\hline
\textbf{Sample 6 with $t = 4$} \\
\hline
$s_4^t$ & 4.6  units \\
$c_4^t$ & 6.1  units \\
$e_4^t$ & 26.4 units \\
\hline

\end{tabular}
\caption{\label{tab:geru}
Euclidean Distance between the same representations for Sentiment Regression as compared to Emotion Detection. (Distances have been converted to units for convenience and easier comparison)
}
\end{table}

\section{Conclusion}
In this paper, we have presented an RNN architecture for multi-modal sentiment analysis and emotion detection in conversation. In contrast to the current state-of-the-art models, our model focuses on effectively capturing the context of a conversation and treats each modality independently, taking into account the information a particular modality is capable of holding. Our model consistently performs well on benchmark datasets such as CMU-MOSI and CMU-MOSEI in any multi-modal setting.

The model can be further extended to have better feature extractors, and increase both the number of modalities and the number of participants in the conversation. Due to the lack of availability of datasets consisting of these extensions with emotion or sentiment labels, we have left this to our future work.

\bibliography{acl2020}
\bibliographystyle{acl_natbib}

\end{document}